%% file: main.tex
\documentclass[lettersize,journal]{IEEEtran}
\usepackage{amsmath,amsfonts}
\usepackage{algorithmic}
\usepackage{array}
\usepackage[caption=false,font=normalsize,labelfont=sf,textfont=sf]{subfig}
\usepackage{textcomp}
\usepackage{stfloats}
\usepackage{url}
\usepackage{verbatim}
\usepackage{graphicx}
\usepackage{cite}
\usepackage{microtype}
\usepackage{booktabs}
\usepackage{indentfirst}
\usepackage{adjustbox}
\usepackage{tabularx}
\usepackage{color}
\usepackage{makecell}
\usepackage{xcolor,colortbl}
\usepackage[english]{babel}
\usepackage{multirow}
\usepackage{tikz}
\usetikzlibrary{spy,backgrounds}
\usepackage{soul}
\usepackage{subcaption}
\usepackage{enumitem}
\usepackage{dsfont}

\usepackage{amssymb}
\usepackage{hyperref}
\usepackage{arydshln}
\usepackage{float}
\usepackage{amssymb}
\usepackage{tikz}
\usetikzlibrary{arrows,chains,matrix,positioning,scopes}
\usepackage{booktabs}
\usepackage{lipsum}
\usepackage[ruled]{algorithm2e}

\newcommand{\tabincell}[2]{\begin{tabular}
		{@{}#1@{}}#2\end{tabular}}

\newcommand{\name}{ROS-GS\xspace}
\newcommand{\lgs}{3D Gaussian Splatting\xspace}
\newcommand{\sgs}{3DGS\xspace}
\newcommand{\ltwodgs}{2D Gaussian Splatting\xspace}
\newcommand{\stwodgs}{2DGS\xspace}

\begin{document}

\title{\name: Relightable Outdoor Scenes With\\ Gaussian Splatting}

\author{
Lianjun Liao, 
Chunhui Zhang,
Tong Wu,
Henglei Lv,
Bailin Deng
and 
Lin Gao

\thanks{Corresponding author is Lin Gao (gaolin@ict.ac.cn).}
\thanks{Lianjun Liao and Chunhui Zhang are with the School of Information Science and Technology, North China University of Technology, Beijing, China. Email: LiaoLianjun@ncut.edu.cn, harukizhang@mail.ncut.edu.cn.}
\thanks{Tong Wu, Henglei Lv and Lin Gao are with the Beijing Key Laboratory of Mobile Computing and Pervasive Device, Institute of Computing Technology, Chinese Academy of Sciences, Beijing, China, and also with the University of Chinese Academy of Sciences, Beijing, China. E-Mail: wutong0317@outlook.com, \{lvhenglei22s, gaolin\}@ict.ac.cn.}
\thanks{Bailin Deng is with the School of Computer Science and Informatics, Cardiff University, Cardiff, United Kingdom. Email: DengB3@cardiff.ac.uk.}
}

\markboth{~}
{~}

\IEEEpubid{~}

\maketitle

\input{0_abstract}

\begin{IEEEkeywords}
Gaussian splatting, inverse rendering, relighting.
\end{IEEEkeywords}

\input{1_intro}

\input{2_related}
\input{3_method}

\input{4_results}

\input{5_conclusion}

\bibliographystyle{IEEEtran}
\bibliography{main}

\end{document}

%% file: 0_abstract.tex
\begin{abstract}
Image data captured outdoors often exhibit unbounded scenes and unconstrained, varying lighting conditions, making it challenging to decompose them into geometry, reflectance, and illumination.
Recent works have focused on achieving this decomposition using Neural Radiance Fields (NeRF) or the \lgs~(\sgs) representation but remain hindered by two key limitations: the high computational overhead associated with neural networks of NeRF and the use of low-frequency lighting representations, which often result in inefficient rendering and suboptimal relighting accuracy.
We propose \name, a two-stage pipeline designed to efficiently reconstruct relightable outdoor scenes using the Gaussian Splatting representation. 
By leveraging monocular normal priors, \name first reconstructs the scene's geometry with the compact \ltwodgs~(\stwodgs) representation, providing an efficient and accurate geometric foundation. 
Building upon this reconstructed geometry, \name then decomposes the scene's texture and lighting through a hybrid lighting model. 
This model effectively represents typical outdoor lighting by employing a spherical Gaussian function to capture the directional, high-frequency components of sunlight, while learning a radiance transfer function via Spherical Harmonic coefficients to model the remaining low-frequency skylight comprehensively.
Both quantitative metrics and qualitative comparisons demonstrate that \name achieves state-of-the-art performance in relighting outdoor scenes and highlight its ability to deliver superior relighting accuracy and rendering efficiency.
\end{abstract}

%% file: 1_intro.tex
\section{Introduction}

\IEEEPARstart{O}{utdoor} relighting plays a crucial role in a wide range of applications, including augmented reality, virtual production, urban planning, and autonomous driving simulation. 
This capability enables the seamless integration of virtual objects with real-world environments and provides realistic visualization of infrastructure and objects under diverse and dynamic lighting scenarios.
For instance, augmented reality applications rely on accurate relighting to maintain consistency between virtual elements and their surroundings, while urban planning and simulation tools require realistic lighting to assess the visual and functional impact of structures under varying conditions.

\begin{figure*}[t]
    \centering
    \includegraphics[width=\linewidth]{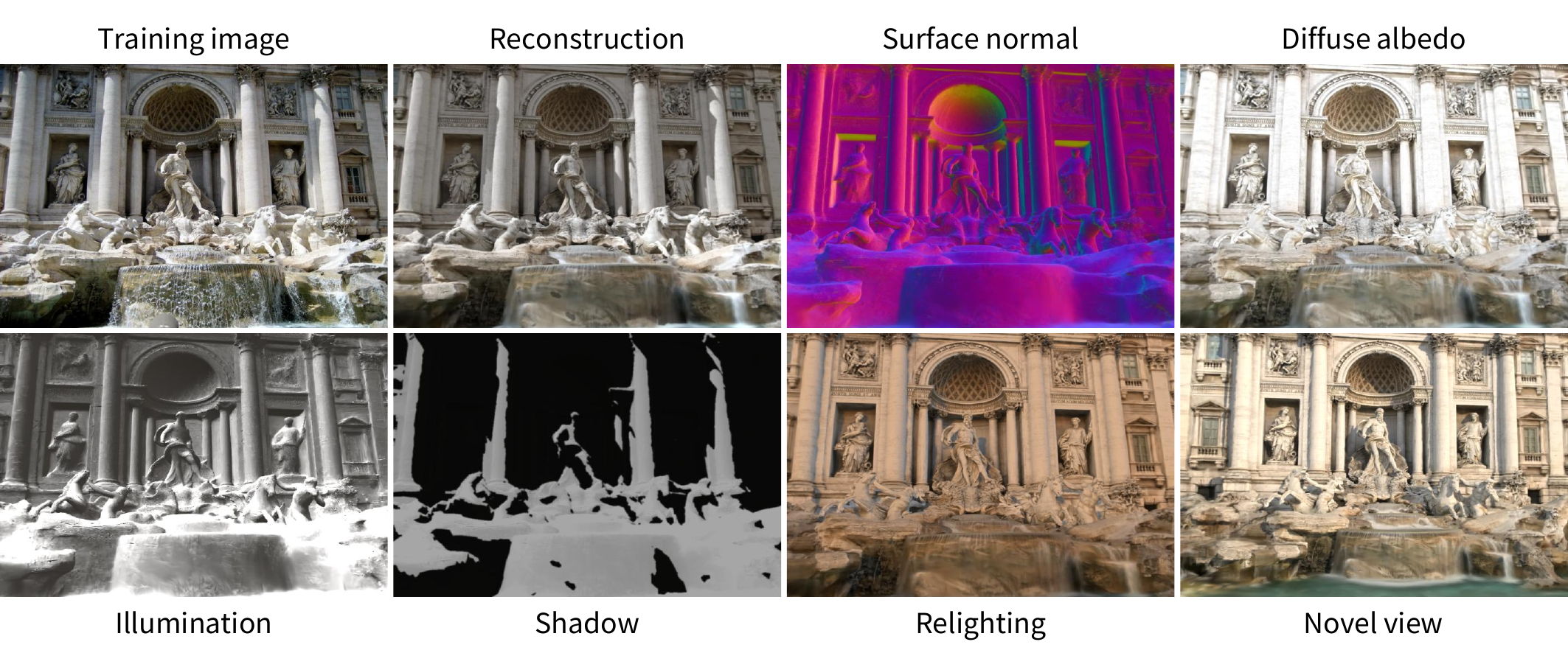}
    \caption{Given multi-view images with unconstrained lighting conditions, \name optimizes its Gaussian Splatting representation and decomposes the geometry, texture, and lighting components of the outdoor scene, which enables not only faithful novel view synthesis but also realistic and accurate relighting.}
    \label{fig:teaser}
\end{figure*}

Despite its significance, outdoor relighting remains a challenging problem due to the inherent complexity of natural lighting and scene interactions. 
Outdoor environments experience continuously changing illumination conditions dictated by the sun's position, weather variations, and atmospheric phenomena such as light scattering and absorption. 
These dynamic factors produce intricate illumination patterns, making it difficult to accurately replicate unknown lighting conditions from captured outdoor scenes. 
Furthermore, the vast scale of outdoor scenes, with their intricate details and wide-ranging features, poses a challenge in balancing efficiency and realism. 
\IEEEpubidadjcol
Addressing these challenges requires techniques capable of handling multiple unknown lighting conditions and the large-scale, computationally intensive nature of outdoor relighting, paving the way for advancements across multiple domains.

Current state-of-the-art methods~\cite{NeRF-OSR,solnerf,NeuSky,SRTensoRF} leverage neural implicit representations like Neural Radiance Fields (NeRF)~\cite{NeRF}, which encode scene geometry and appearance using neural networks. 
These methods have shown promising results in relighting by modeling light interactions and producing high-quality visual outputs.
While they can handle multiple unknown lighting conditions, they fall short in efficiency because of costly volume rendering in NeRF. 
Such inefficiency is particularly problematic for outdoor environments, where large-scale scenes and dynamic lighting demand both speed and flexibility.

Recently proposed Gaussian Splatting-based methods have improved efficiency largely.
However, works like Relightable 3D Gaussians~\cite{R3DG,GShader,IRGS} are designed for single object relighting and can only handle a single unknown lighting condition.
ReCap~\cite{ReCap} can model multiple lighting conditions but only focus on the synthetic single object and does not account for self-shadowing effects.
There are other works that aim to relight scenes, however, among them, GS-IR~\cite{GS-IR} is constrained to single lighting condition, and methods like $GS^3$~\cite{TripleSplat,RNG} are restricted to a known one-light-at-a-time (OLAT) setting. 
Apart from above, LumiGauss~\cite{LumiGauss} aims to relight outdoor scenes with multiple unknown lighting, but its use of a low-frequency lighting representation solely based on Spherical Harmonics prevents it from accurately modeling high-frequency illumination effects, such as cast shadows and high-contrast light-dark boundaries, which are essential for outdoor realistic relighting. 
These limitations highlight the need for more advanced methods that integrate computational and storage efficiency while effectively handling multiple unknown illumination conditions in outdoor captures. 
Additionally, such techniques must enable the accurate representation of fine-grained lighting interactions in outdoor scenes.

To address these challenges, we propose a novel framework for outdoor relighting that leverages the fast rendering speed of \ltwodgs and the realism of physically-based rendering to achieve photorealistic lighting manipulation for outdoor scenes.
Our approach comprises two stages inspired by previous inverse rendering methods~\cite{solnerf,NeRFactor}. 
In the first stage, we incorporate the compact \stwodgs representation~\cite{2dgs} and monocular normal estimation prior~\cite{StableNormal} to reconstruct outdoor scenes' geometry from a set of unconstrained images. 
Then in the second stage, based on the reconstructed geometry, we introduce physically-based rendering into the \stwodgs's rendering process, where the Spherical Harmonics attribute of Gaussians is replaced with view-independent texture parameters. 
In addition, inspired by SOL-NeRF~\cite{solnerf}, a hybrid lighting representation is adopted for modeling high-frequency sun light and low-frequency sky light.
By progressively optimizing texture and lighting, \name can decompose the geometry, texture, and lighting from multi-view images, making outdoor relighting and real-time rendering plausible. 
The contribution of our method can be summarized as follows:
\begin{itemize}[leftmargin=*]
    \item{\name leverages the \stwodgs representation and the monocular normal estimation into the inverse rendering pipeline, producing more faithful geometry and texture estimation.}
    \item{By incorporating the hybrid lighting representation with the Gaussian Splatting representation, \name enables realistic relighting and real-time dynamic shadow synthesis.}
    \item{Both qualitative and quantitative results show that \name outperforms baseline methods in outdoor scene decomposition and relighting tasks.}
\end{itemize}

%% file: 2_related.tex
\section{Related Works}

\subsection{Novel View Synthesis and Geometry Reconstruction}
Novel view synthesis has long been a core topic in computer graphics and vision. 
Early works used light field interpolation~\cite{LFR,ULF}, while later methods employed depth warping~\cite{Song19,SynSin} and Multi-Plane Images (MPIs)\cite{3DPhoto,HanWY22,WangLSSCK22,LuvizonCSCFDSPJ21}, though these were limited to small camera movements. 
More advanced representations, such as meshes, voxels, implicit representation~\cite{IMNet,DeepSDF,OccNet,DVR,NeRF} and neural rendering~\cite{DeferredNR,TextureFields}, enabled greater flexibility but often suffered from rendering inefficiencies. 
Recently, \sgs~\cite{3dgs} emerged, efficiently modeling scenes as Gaussian ellipsoids for fast and high-quality rendering.

Traditional methods use hand-crafted features to match pixels across views, reconstructing surface from extracted point clouds with Poisson surface reconstruction~\cite{PoissonSR,SPoissonSR} or Marching Cubes~\cite{MC}. 
With the advent of Neural Radiance Fields (NeRF)~\cite{NeRF} enabling photorealistic novel view synthesis, recent works have leveraged NeRF for geometry reconstruction. 
NeuS~\cite{NeuS} introduced an unbiased, occlusion-aware formulation connecting signed distance fields to volumetric rendering, while 
UNISURF~\cite{UNISURF} linked occupancy networks with NeRF. 
Building on NeuS, SparseNeuS~\cite{SparseNeuS} enabled sparse-view reconstruction, and 
MonoSDF~\cite{MonoSDF} improved geometry quality using monocular normal estimation. 
With \sgs improving the rendering speed, SuGaR~\cite{SuGaR} introduces a self-regularization term for better surface alignment. 
\stwodgs~\cite{2dgs} and GS-Surfel~\cite{gssurfel} introduces more compact 2D Gaussian representation for better surface reconstruction.

\subsection{Relighting}
Early relighting works usually capture scenes under varying illumination~\cite{BiXSHHKR20,Bi2008} to estimate material properties and environmental lighting for relighting purposes. 
With neural rendering enabling realistic novel view synthesis, works start to recover geometry, texture, and lighting from casually captured multi-view images. The pioneering work PhySG~\cite{PhySG} adopts signed distance function as geometry representation and uses spherical Gaussians (SG) to approximate lighting.
NDR~\cite{NDR} and NDRMC~\cite{NDRMC} incorporate the deformable tetrahedra representation~\cite{DMTet} for efficient geometry extraction and rendering. 
NeRFactor~\cite{NeRFactor} further introduces prior knowledge from existing material datasets for more accurate texture and lighting decomposition. 
For more complex materials, implicit lighting representation~\cite{NeRO,DE-NeRF,VD-NeRF,RefNeRF} is proposed to better capture high-frequency lighting information. 
Regarding the outdoor scene setting, NeRF-OSR~\cite{NeRF-OSR} utilizes a Spherical Harmonics~(SH) function to represent lighting information to approximate visibility with a neural network. 
SOL-NeRF~\cite{solnerf} instead introduces a hybrid lighting formulation, where a spherical Gaussian~(SG) function models the sun light and a Spherical Harmonics~(SH) function represents the sky light, which makes creating cast shadow possible.
To improve the shadow calculation efficiency, NeuSky~\cite{NeuSky} proposes a spherical directional distance function to approximate the visibility information among the scene. 

More recently, \sgs-based relighting methods have emerged. 
Relightable 3D Gaussians~\cite{R3DG} incorporate a ray-tracer to 3DGS in order to depict shadow effects. While this method can achieve continuously moving shadow effects, it fails to produce accurate self-shadows with sharp edges.
Similarly, IRGS~\cite{IRGS} implements shadow effects using a 2D Gaussian ray-tracer. However, it focuses solely on object relighting rather than whole-scene relighting and can only handle a single unknown lighting condition.
GS$^3$~\cite{TripleSplat} focused on single objects under known OLAT lighting condition. 
Recap~\cite{ReCap} aimed to improve inverse rendering using multiple lighting conditions but was confined to synthetic objects, neglecting secondary effects like shadows. Its use of a learnable image-based cubemap for lighting is memory-intensive and computationally costly, proving inefficient for the numerous lighting conditions typical in outdoor captures.
LumiGauss~\cite{LumiGauss} integrates Precomputed Radiance Transfer (PRT) with 2DGS for outdoor scenes, but its omnidirectional lighting assumption limits high-frequency effects such as sharp shadows, and PRT overfitting often results in baked-in artifacts.
Notably, the concurrent work GaRe~\cite{GaRe} also tackles unconstrained lighting input, decomposes illumination into sun, sky and indirect component with small MLP mappings, and generates dynamic shadows via a ray-trace-based Gaussian visibility query method. 
In contrast, our method adopts a more compact lighting representation and computes shadows using a mesh-based intersection detection strategy.

\input{fig_overview}

%% file: fig_overview.tex
\begin{figure*}[t!]
    \includegraphics[width=\linewidth]{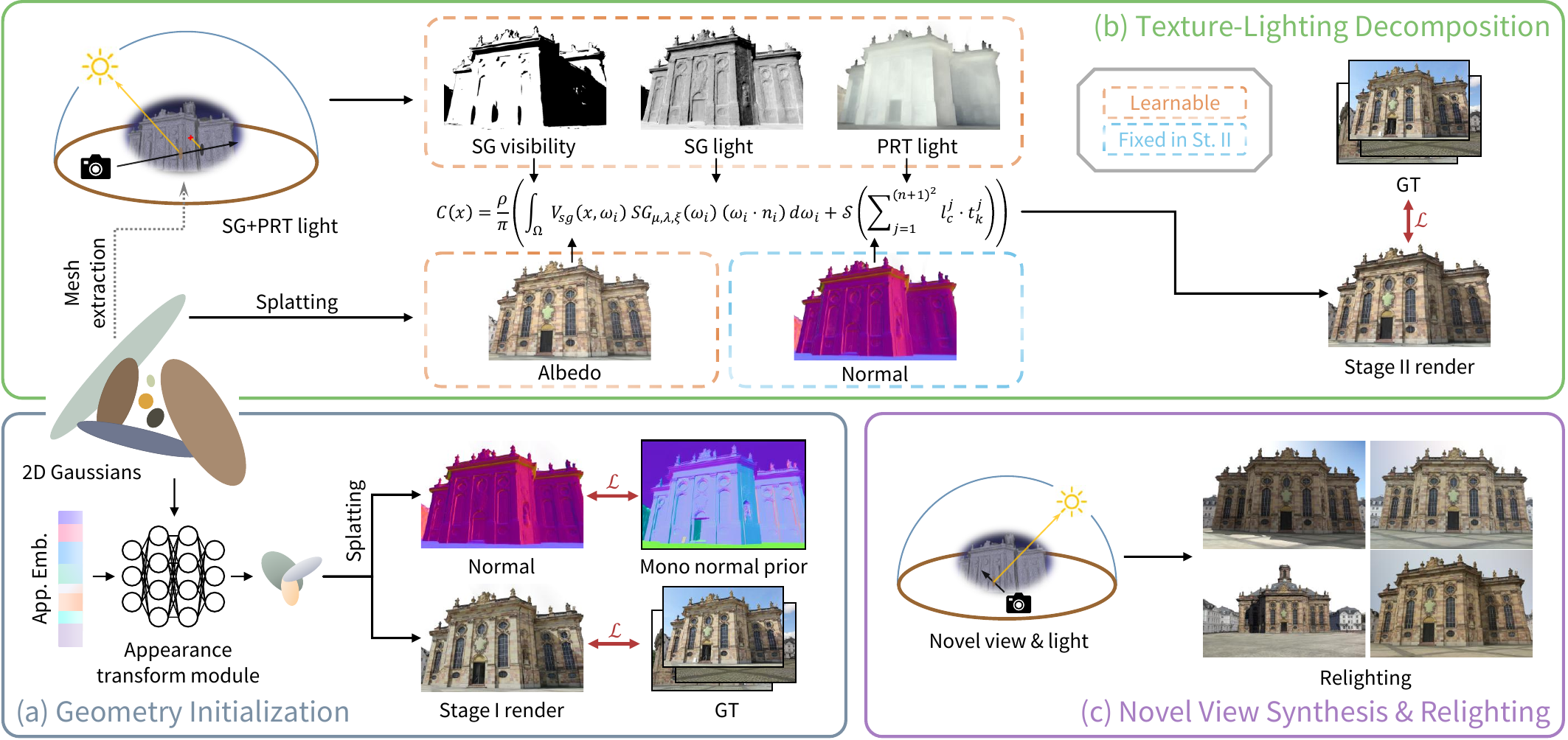}
    \caption{
    \textbf{Pipeline of our method.}
    Given multi-view images captured under unconstrained lighting, our \name introduces a two-stage relighting pipeline based on Gaussian splatting.
    (a) In the first stage, we reconstruct scene geometry using the 2DGS representation with an appearance transformation module, guided by monocular normal priors.
    (b) In the second stage, we employ a hybrid lighting representation to achieve texture–lighting decomposition.
    (c) After training, our model supports both novel view synthesis and realistic relighting of outdoor scenes.
    }
    \label{fig:overview}
\end{figure*}

%% file: 3_method.tex
\section{Method}
\label{sec:method}
The overall pipeline of our method is illustrated in Fig.~\ref{fig:overview}. 
Given multi-view images of outdoor scenes captured under unconstrained lighting conditions, \name decouples geometry reconstruction from texture-lighting decomposition through a two-stage design.
This decomposition subsequently enables realistic relighting of the scenes using the Gaussian Splatting representation~\cite{3dgs}. 
The first stage focuses on reconstructing the scene geometry utilizing the compact 2DGS representation~\cite{2dgs} combined with an appearance transformation module, guided by monocular normal estimation priors.
Afterward, the second stage introduces a hybrid sun-sky lighting model to effectively disentangle texture and illumination, enabling high-quality relighting with faithful shadow effects.

\subsection{Preliminaries}
\label{preliminaries}

\subsubsection{2D Gaussian Splatting}
\name utilizes 2D Gaussian Splatting~\cite{2dgs}, a compact geometric representation capable of reconstructing scene surfaces with smooth normals from multi-view images. A 2D Gaussian surfel is defined by:
\begin{equation}
\mathcal{G}(p)=\exp{(-\frac{1}{2} (p - p_k)^{\top} \Sigma^{-1} (p-p_k))},
\end{equation}
where $p_k$ is the center of a Gaussian surfel, and $\Sigma=RSS^{\top}R^{\top}$ is its covariance matrix, parameterized by a scaling matrix $S$ and a rotation matrix $R$.
In the rendering process, Gaussian surfels in world space are first transformed into camera coordinates. They are then projected onto the image plane using an explicit ray-splat intersection technique, which facilitates stable optimization.  
Then, a volumetric alpha blending process is employed to compute the color of a pixel by integrating contributions from front to back:
\begin{equation}
\label{eqn:gs_rendering}
c(\mathbf{x})=\sum\nolimits_{i=1}{c_i \alpha_i \hat{\mathcal{G}}_i(\mathbf{u}(\mathbf{x}))} \prod\nolimits_{j=1}^{i-1}{(1 - \alpha_j \hat{\mathcal{G}}_j(\mathbf{u}(\mathbf{x})))},
\end{equation}
where $c_i$ and $\alpha_i$ are the view-dependent appearance and the alpha value of the $i$-th Gaussian. 
$\hat{\mathcal{G}}_k(\mathbf{u}(\mathbf{x}))$ ($k$=$i$,$j$) is the projected 2D Gaussian function, determined by its projected covariance matrix $\Sigma'$ evaluated at the intersection point 
$\mathbf{u}(\mathbf{x})$ between a camera ray and the Gaussian. 
$c(\mathbf{x})$ is the alpha-blended color along the ray $\mathbf{x}$.

\subsubsection{Rendering Equation}
To achieve relighting under novel illumination, it is crucial to model texture and lighting independently. 
\name incorporates physically-based rendering principles within the Gaussian Splatting framework, guided by the rendering equation~\cite{kajiya_rendering}:
\begin{equation}\label{equation::rendering_equation}
    L(x,\omega_o) = \int_{\Omega} f_{r}(x, \omega_o, \omega_i)  L_{in}(x, \omega_i)  (\omega_i \cdot n ) \, \mathrm{d} \omega_i.
\end{equation}
Here, $x$ denotes a point on a surface with normal $n$. 
$L_{in}(x, \omega_i)$  is the incident lighting at point $x$ from direction $\omega_i$. 
The function $f_{r}(x, \omega_o, \omega_i)$ represents the BRDF at point $x$, which depends on both the light direction $\omega_i$ and the viewing direction $\omega_o$. 
The outgoing radiance $L(x, \omega_o)$ is obtained by integrating these contributions over the hemisphere $\Omega= \{ \omega_i: \omega_i \cdot n > 0 \}$.

\subsubsection{Precomputed Radiance Transfer}
Sloan et al.~\cite{sloan2002precomputed} propose that the rendering equation can be viewed as an integral over the product of the incident illumination $L_{in}(x, \omega_i)$ and the radiance transfer function $T(x,\omega_i, \omega_o)=f_{r}(x, \omega_o, \omega_i)(\omega_i \cdot n)$.
Using Spherical Harmonics (SH) basis functions approximates low-frequency illumination efficiently, as the integral reduces to a simple dot product of SH coefficients due to their orthonormality. 
For diffuse surfaces, the radiance transfer simplifies to a vector $T(x)$, yielding the outgoing light $L(x)$ as:
\begin{equation}
    L(x) = \sum\nolimits_{k=1}^{(n+1)^2} T(x)^k L_{in}^k,
\end{equation}
where $T(x)^k$ and $L_{in}^k$ are the $k$-th SH coefficients of the transfer function and the incident lighting, respectively.

\subsection{Geometry Initialization} \label{method_stage_i}
\name takes as input $N$ multi-view images of an outdoor scene under unconstrained lighting conditions.
The first stage of our pipeline focuses on geometry initialization.
Reconstructing accurate geometry from such inputs is challenging due to complex appearance variations across views and weakly textured regions, often further exacerbated by hard shadows.

To address these challenges, we integrate an appearance transformation module, inspired by NeRF-in-the-Wild~\cite{NeRFintheWild} and WildGaussians~\cite{WildGaussians}, into the standard 2DGS training.
In this stage, each Gaussian is defined as:
\begin{equation}
    \mathcal{G}_I=\{p_k, \sigma_k, R_k, o_k, \rho_k, \Lambda_k \}, k\in K,
\end{equation}
where $p_k$, $\sigma_k$, $R_k$, and $o_k$ denote the position, scaling, rotation, and opacity of the $k$-th Gaussian, respectively.
$\rho_k \in [0, 1]^3$ denotes the base color of the $k$-th Gaussian, which provides a simpler and more compact representation than using Spherical Harmonics (SH).
To model appearance variability, we introduce a per-Gaussian embedding $\Lambda_k$ that captures local appearance variations, and a set of per-image embeddings $\{\Xi_j\}_{j=1}^{N}$ that accounts for lighting changes across different views.
The appearance transformation module is an MLP $f_{app}$ that predicts the parameters of an affine transformation:
\begin{equation}
    (\gamma, \beta) = f_{app}( \rho_k, \Lambda_k, \Xi_j ),
\end{equation}
which adjusts the base color $\rho_k$ to produce the transformed appearance color $\tilde{\rho}_k$:
\begin{equation}
    \tilde{\rho}_k=\gamma \cdot \rho_k + \beta.
\end{equation}
The resulting appearance color $\tilde{\rho}_k$ is then rasterized using Eq.~\eqref{eqn:gs_rendering}.

Although this formulation improves robustness, weakly textured and planar regions still hinder the optimization of smooth surface normals.
To further enhance geometry quality, we incorporate a monocular normal prior.
Since our inputs are captured under unconstrained conditions, moving foreground objects often cause multi-view inconsistencies.
To ensure stable supervision, we identify these dynamic regions and exclude them using semantic masks generated by an off-the-shelf segmentation method~\cite{MMSegmentation}.
In addition, we apply the normal prior only to non-edge regions. Specifically, we define non-edge regions as image areas that do not correspond to geometric discontinuities. To extract them, we use the classical Canny edge detector with thresholds set to [100, 200], followed by dilation with a 3×3 kernel.
We focus on non-edge regions because planar areas benefit most from normal constraints, whereas applying the prior across the entire image leads to over-smoothed normals.

Finally, to prepare for shadow computation in the second stage, we extract a mesh from the reconstructed geometry. 
We first render depth maps from the training views, filter out sky, foliage, and other dynamic occluders using semantic masks, and then fuse the depth maps via the truncated signed distance function (TSDF) integration method to obtain clean surface meshes.

\subsection{Texture-Lighting Decomposition} \label{method_stage_ii}
In the second stage, building upon the geometry initialized in the first stage, \name focuses on decomposing the scene's texture and illumination. 
We employ a hybrid lighting representation, inspired by SOL-NeRF~\cite{solnerf}, to model the complex lighting information present in outdoor scenes.
Specifically, we employ a single spherical Gaussian (SG) function to capture the directional, high-frequency components of sunlight. 
For the modeling of the skylight, SOL-NeRF employs a single set of SH functions combined with an MLP-predicted ambient occlusion term.
However, this introduces significant computational overhead during inference.
In contrast, our approach leverages Precomputed Radiance Transfer (PRT) to model skylight in a more comprehensive and spatially adaptive manner. By decoupling lighting transport from geometry and precomputing visibility interactions, our method achieves more realistic and detailed illumination without relying on expensive per-pixel MLP evaluations.
We define globally shared SH coefficients $l_c \in \mathbb{R}^{3 \times (n+1)^2}$ to represent the low-frequency ambient illumination, where $c=1,\ldots,N$ indexing the input views. For a balance between efficiency and quality, we adopt second-order SH ($n=2$)~\cite{Ravi2001}.
Meanwhile, the per-Gaussian SH coefficients $t_k \in \mathbb{R}^{1 \times(n+1)^2}$ encode the radiance transfer function.
Thus, each Gaussian in this stage is defined as:
\begin{equation}
    \mathcal{G}_{II}=\{p_k, \sigma_k, R_k, o_k, \rho_k, t_k \}, k\in K,
\end{equation}
where $p_k$, $\sigma_k$, $R_k$ and $o_k$, as geometry attributes of Gaussians, are all inherited from the first-stage reconstruction. 
For $\rho_k \in [0,1]^3$, we reuse the same parameter vector from the first stage, but reinterpret it as the albedo of the $k$-th Gaussian in the second stage.

Assuming the sun and sky are infinitely distant, incoming light rays can be treated as parallel, making the incident illumination primarily dependent on surface orientation rather than spatial position. Under this assumption, accurate and smooth surface normals are crucial for faithful estimation of lighting.
While the final accumulated normals produced by \ltwodgs are generally smooth, the local normals of individual Gaussians contributing to a ray can exhibit significant variance. To ensure more consistent multi-view geometry and shading, we adopt a deferred shading technique~\cite{DeferredGS,yeDeferred}, performing shading calculations in screen space at each pixel using the final accumulated normals rather than on a per-Gaussian basis.
Besides, we primarily focus on diffuse surfaces and replace the BRDF $f_{r}(x, \omega_o, \omega_i)$ with a simplified diffuse model $f_{d}= \frac{\rho}{\pi}$ where ${\rho}$ is the diffuse albedo, which also makes the outgoing light independent of the viewing direction. 
Thus, for each pixel $x$, the rendered color is computed as:
\begin{equation}
    C(x)=\frac{\rho}{\pi} (I_{sun}+I_{amb}).
\end{equation}
Here, $\rho$ is an albedo map generated by splatting the albedo attribute $\rho_k$ of Gaussians, $I_{sun}$ is the irradiance contribution from the sun represented by an SG, and $I_{amb}$ is the irradiance from the ambient skylight represented by SH.

A spherical Gaussian is defined as $SG_{\mu, \lambda, \xi}(\nu) = \mu \mathrm{e}^{\lambda(\nu \cdot \xi - 1)}$, where $\mu \in \mathbb{R}_+^3$ is the lobe amplitude, $\lambda \in \mathbb{R}_+$ controls the lobe sharpness, $\xi \in \mathbb{S}^2$ is a unit vector representing the sun direction, and $\nu$ is the input direction.
The irradiance of the sunlight component is then computed as:
\begin{equation}
    I_{sun}=\int_{\Omega} V_{sun}(x,\omega_i) SG_{\mu, \lambda, \xi}(\omega_i) (\omega_i \cdot n ) \, \mathrm{d} \omega_i,
\end{equation}
where, $V_{sun}(x, \omega_i) \in \{ 0, 1 \} $ denotes the visibility of the surface point $\mathbf{X}$ with respect to the incident direction $\omega_i$, $\mathbf{X}$ is the surface point corresponding to the pixel $x$, and $n$ denotes the normal map.
Visibility is calculated for each pixel in screen space in a deferred manner. We perform ray tracing against the mesh extracted in the first stage, accelerated by a Bounding Volume Hierarchy (BVH) tree. $V=1$ if the light from direction $\omega_i$ reaches the point $x$, and $V=0$ if the ray intersects the mesh.

For the ambient light irradiance $I_{amb}$, we first compute the ambient irradiance at each Gaussian by the dot product between the global and local SH coefficients, and then splat these per-Gaussian ambient contributions to produce the ambient irradiance map:
\begin{equation}
    I_{amb}=\mathcal{S}(\sum\nolimits_{j=1}^{(n+1)^2} l_c^j \cdot t_{k}^j),
\end{equation}
where $\mathcal{S}(\cdot)$ denotes the 2DGS rasterization process for splatting Gaussian attributes.

\subsection{Training losses} \label{train_losses}
Our model utilizes the 2DGS rasterization process for image-based scene reconstruction. The training is divided into two stages, each with a specific loss function.
The loss function for the first stage, focused on geometry reconstruction, is defined as:
\begin{equation}
    \mathcal{L}_{1st}=\lambda_1\mathcal{L}_{render}+\lambda_2\mathcal{L}_{reg}
    +\lambda_3\mathcal{L}_{np}+\lambda_4\mathcal{L}_{mask}.
\end{equation}
Here, $\mathcal{L}_{render}$ denotes a rendering loss that measures the difference between the rendered images and the ground truth images, and is composed of an $L_1$ term and a D-SSIM~\cite{dssim} term. 
$\mathcal{L}_{reg}$ is a term introduced in 2DGS for normal consistency and distortion regularization and retained in our pipeline.
$\mathcal{L}_{np}$ and $\mathcal{L}_{mask}$ are geometry prior losses for normal prior and semantic mask, respectively.

For $\mathcal{L}_{np}$, we obtain a prior normal map $n_p$ from an off-the-shelf monocular normal estimator~\cite{StableNormal}, and compare it against both the alpha-blended rendered normal $\hat{n}_r$ and the normal $\hat{n}_s$ derived from the depth map gradient via:
\begin{equation}
    \mathcal{L}_{np}= (\sum{(1-n_{p}^{\top} \hat{n}_{r})} +  \sum{( 1-n_{p}^{\top} \hat{n}_{s})}).
\end{equation}
This ensures consistency between the geometries defined by depth and normals.

For $\mathcal{L}_{mask}$, we generate semantic masks for the input images using MMSegmentation~\cite{MMSegmentation}, and compute a binary cross entropy loss between the sky regions in the masks and the rendered alpha channel. 
Additionally, we use the semantic masks to filter out dynamic object regions when computing the rendering loss and the normal prior loss, thereby focusing the training on the static primary components of the scene.

In the second stage, we fix the geometry reconstructed in the first stage and focus on texture and lighting decomposition. The loss function is defined as:
\begin{equation}
    \mathcal{L}_{2nd}=\lambda_1\mathcal{L}_{render}+\lambda_5\mathcal{L}_{sp}+\lambda_6\mathcal{L}_{amb},
\end{equation}
where we employ two loss terms $\mathcal{L}_{sp}$ and $\mathcal{L}_{amb}$ for priors on the sunlight color and ambient light, respectively.

The sunlight color prior $\mathcal{L}_{sp}$ constrains the intensity of the SG lobe within the typical color range of sunlight \cite{solnerf}. This prior is formulated as a piecewise polynomial function $f_{sun}$ that fits the result values of the modified Nishita model~\cite{nishita_sunlight_1993} for training efficiency, and the term is defined as:
\begin{equation}
    \mathcal{L}_{sp} = \lVert f_{sun}(\theta) - \mu \rVert_{1},
\end{equation}
where $\mu$ denotes the learned SG intensity and $\theta$ is a simulated solar elevation angle parameter computed from the sun direction $\xi$ in our SG lighting model.

The ambient light loss $\mathcal{L}_{amb}$ encourages the ambient SH component to capture low-frequency illumination and prevents overfitting hard-edged baked-in shadows, and is defined as the total variation of the rendered ambient irradiance map $I_{amb}$:
\begin{equation}
    \begin{split}
        \mathcal{L}_{amb} = \sum\nolimits_{x} \Big(
            & \lVert I_{amb}(x+\mathbf{e}_x) - I_{amb}(x) \rVert_1 \\
            & + \lVert I_{amb}(x+\mathbf{e}_y) - I_{amb}(x) \rVert_1
        \Big),
    \end{split}
\end{equation}
where $\mathbf{e}_x$, $\mathbf{e}_y$ denote unit pixel shifts along the horizontal and vertical directions.

%% file: 4_results.tex
\section{Experiments and Results}
\subsection{Implementation Details}

For evaluation, we conduct experiments and ablation studies on two datasets: a synthetic dataset from SOL-NeRF~\cite{solnerf} and a real dataset from NeRF-OSR~\cite{NeRF-OSR}. 
Since the NeRF-OSR dataset does not provide ground-truth texture or normal maps, we report quantitative decomposition results only on the three synthetic scenes from the SOL-NeRF dataset.
For quantitative relighting evaluation, we follow prior works and report results on the three scenes of the NeRF-OSR dataset.
We adopt commonly used metrics for decomposition and relighting evaluation, including PSNR, SSIM~\cite{SSIM}, and Mean Squared Error (MSE). 
For geometry reconstruction evaluation, we use the Mean Absolute Error (MAE) between the decomposed normal maps and the ground truth normal maps.

We run all experiments using a single NVIDIA RTX 3090 GPU with 24GB VRAM.
We set $\lambda_1=1.0$ for both $L_1$ loss and D-SSIM loss for rendered images and set $\lambda_2=0.05$ for both normal consistency loss and distortion loss.
We set $\lambda_3=0.1$ for the normal prior loss $\mathcal{L}_{np}$ and set $\lambda_4=0.1$ for the semantic mask loss $\mathcal{L}_{mask}$.
In addition, we set $\lambda_5=10.0$, decaying to 0.01 over 50k iterations, for the sunlight color prior loss. This ensures that $\mathcal{L}_{sp}$ acts as a strong constraint in the early phase, stabilizing sunlight color learning.
Finally, we set $\lambda_6=1.0$ for the loss $\mathcal{L}_{amb}$ on the ambient irradiance map.
For appearance transformation, we use embeddings of size 24 for the per-Gaussian embedding $\Lambda_k$ and 32 for the per-image embedding $\Xi_j$. 
For the appearance MLP $f_{app}$, we use one hidden layer of size 128.

We employ a progressive training strategy to effectively decompose scene geometry, appearance, and illumination. In the first stage, optimization is restricted to scene geometry. During the second stage, all geometry-related Gaussian attributes are frozen. Since the appearance transformation module in the first stage tends to overfit the scene appearance, the model does not inherit a well-defined albedo at the beginning of the second stage. Thus, during the initial 10,000 iterations of the second stage, shading is omitted to allow the model to learn an initial, refined albedo. For the following 40,000 iterations, all Gaussian attributes, including the newly learned albedo, are frozen to focus exclusively on optimizing the lighting parameters. Finally, the last 50,000 iterations involve the joint optimization of non-geometric Gaussian attributes and the lighting parameters, enabling all components to be fine-tuned together.

\input{table_recon}

\input{fig_decomp}

\input{fig_decomp_extra}
\input{fig_decomp_neusky}
\input{table_relight}
\input{fig_relight}
\input{fig_shadow_moving}

\subsection{Decomposition Results}

\name performs decomposition of geometry, texture, and lighting from multi-view images. 
We compare the decomposition results of our approach with four baseline methods: LumiGauss~\cite{LumiGauss}, ReCap~\cite{ReCap}, NeRF-OSR~\cite{NeRF-OSR}, and SOL-NeRF~\cite{solnerf}.
We present qualitative results on three real scenes from the NeRF-OSR dataset in Fig.\ref{fig:decomp} and on two synthetic scenes from SOL-NeRF in Fig.\ref{fig:decomp_extra}.
We further compare with NeuSky~\cite{NeuSky} on a real scene in Fig.\ref{fig:decomp_neusky}.
We report quantitative results on synthetic scenes from SOL-NeRF in Table~\ref{tab:scene_decomp}.
For geometry recovery, our approach produces normal maps with clearer structural details and smoother surfaces compared to the baseline methods. 
In terms of appearance decomposition, our decomposed albedo maps not only exhibit more natural colors but also contain fewer shadow artifacts, unlike those produced by LumiGauss and ReCap, which often contain baked-in shadows. 
Moreover, our approach demonstrates superior illumination expressiveness, as evidenced by the enhanced quality of reconstruction and the more accurately decoupled shadow maps. Compared to baseline methods, our model can handle high-frequency illumination effects, such as sharper cast shadows and high-contrast light-dark boundaries.
Overall, our method outperforms these approaches in novel view synthesis and excels in both albedo and normal decomposition, leading to improved scene relighting results, as discussed in later sections.

\subsection{Relighting Results}

We present qualitative relighting results on three real scenes from the NeRF-OSR dataset in Fig.~\ref{fig:relighting}.
For each scene, we show the input image, two relighting results from the same viewpoint, and their corresponding shadow maps. 
We report quantitative results in Table~\ref{tab:relight}.
The shadow maps generated by LumiGauss~\cite{LumiGauss}, which models lighting using low-frequency Spherical Harmonics (SH), fail to capture illumination changes and yield weak cast shadows. 
NeRF-OSR~\cite{NeRF-OSR}, which also adopts SH-based lighting, produces similarly limited shadow effects. 
By contrast, ReCap~\cite{ReCap} does not account for shadows, which results in baked-in shadows within the albedo map and consequently produces incorrect shadow effects. Moreover, its albedo maps also suffer from unnatural color distortions that degrade relighting quality. 
Additionally, the inaccurate and noisy normal maps predicted by LumiGauss result in uneven surface shading and further reduce realism. 
In comparison, our pipeline benefits from monocular geometry estimation, which improves geometric accuracy and supports more reliable decomposition and relighting. 
Moreover, our hybrid lighting formulation produces sharp and realistic shadow effects under varying illumination.
We present a qualitative comparison of shadow movement under changing sunlight directions in Fig.~\ref{fig:shadow_moving}.

\input{fig_ablation_L_np}

\subsection{Efficiency}

Table~\ref{tab:efficiency} compares the training time and FPS of our method with other approaches on the NeRF-OSR dataset. Our training time is comparable to ReCap, and significantly shorter than NeRF-OSR. We also achieve real-time rendering performance that is faster than LumiGauss and significantly outperforms NeRF-OSR.

\subsection{Ablation Study}

In this subsection, we conduct ablation studies to evaluate the impact of several important design choices in our pipeline.

\input{table_efficiency}

\subsubsection{Appearance Transformation Module}
As described in Sec.~\ref{method_stage_i}, we integrate an appearance transformation module into the first stage of our pipeline to handle appearance variations across multi-view inputs.
Its impact is illustrated in the ablation study in Fig.~\ref{fig:ablation_L_np}.
Without its compensation for appearance variations, the estimated normal maps become fuzzy and distorted, thereby hindering the subsequent separation of texture and illumination.
The quantitative results in Table~\ref{tab:ablation} further confirm its effectiveness, demonstrating improved robustness under varying appearance conditions.

\input{table_ablation}

\subsubsection{Normal Prior Loss $\mathcal{L}_{np}$}
During the first stage of geometry reconstruction we introduce a normal prior loss, $\mathcal{L}_{np}$, to encourage more accurate surface normal estimation.
To evaluate its effectiveness, we present an ablation study in Fig.~\ref{fig:ablation_L_np}.
Without the normal prior loss, the normals of large planar regions tend to deviate toward incorrect directions and become distorted, leading to inaccurate decompositions of texture and illumination.
The quantitative results provided in Table~\ref{tab:ablation} further confirm the effectiveness of the normal prior loss, showing clear improvements in the fidelity of the estimated normal maps.

\subsubsection{SH-based Ambient Lighting with AO vs. PRT}
As described in Sec.~\ref{method_stage_ii}, we employ a hybrid lighting representation that combines a spherical Gaussian (SG) sunlight with a PRT-based ambient component.
For the ambient light, we compare our PRT-based model against the alternative approach used in SOL-NeRF~\cite{solnerf}, which represents ambient illumination using Spherical Harmonics (SH) together with an MLP-predicted ambient occlusion (AO).
The qualitative results in Fig.~\ref{fig:ablation_noSG} and the quantitative results in Table~\ref{tab:ablation} demonstrate that our model achieves higher-quality renderings and more accurate albedo estimation.
The shadow map further reveals that the MLP-based AO tends to incorrectly bake material textures into the shadow map.
Moreover, the performance comparison in Table~\ref{tab:efficiency} shows that our model is over three times faster than the SH$\times$AO baseline.

\input{fig_ablation_noSG}

\subsubsection{Lighting Model without SG}
We compare our full model against a baseline that relies solely on an omnidirectional environment light representation.
As shown in Table~\ref{tab:ablation} and Fig.~\ref{fig:ablation_noSG}, our full model produces more accurate and well-decoupled albedo.
In contrast, the baseline without SG tends to overfit illumination, causing shadows to be baked into the albedo or locked into the illumination map, which prevents them from adapting correctly to changes in lighting conditions.

\input{fig_ablation_joint}

\subsubsection{Deferred vs. Forward Shading}
We further conducted an ablation study to evaluate our deferred shading strategy.
In the forward shading variant, shading is computed at each Gaussian before splatting, and shadows are estimated per Gaussian by testing visibility from the SG light to each Gaussian.
As analyzed in Sec.~\ref{method_stage_ii}, while the final alpha-blended normal map appears smooth, the normals at individual Gaussians are often insufficiently accurate for high-quality shading.
As a result, the forward shading approach produces noisy shadow maps and degraded rendering quality, as shown in Fig.~\ref{fig:ablation_joint}.
Quantitative comparisons in Table~\ref{tab:ablation} further confirm the superiority of our deferred shading design.

\subsubsection{Two-Stage vs. Joint Training}
Our method consists of two stages for constructing geometry and for decomposing texture and lighting, respectively.
To assess the effectiveness of this two-stage pipeline, we introduce a baseline that jointly decomposes all three components in a single step.
However, as shown in Fig.~\ref{fig:ablation_joint}, the joint training baseline results in blurry reconstructions and less accurate decomposition.
Further quantitative comparisons in Table~\ref{tab:ablation} confirm the superiority of our two-stage training strategy, highlighting its advantage in achieving more precise and faithful reconstruction and decomposition results.

\input{fig_failure_case}

%% file: table_recon.tex
\begin{table}[tb]
    \caption{
        \textbf{Quantitative comparisons of reconstruction and decomposition results} using SSIM, PSNR, and MSE metrics for rendered image and albedo on the synthetic dataset with baseline methods.
        For the geometry, we compare MAE (Mean Absolute Error) between the rendered and ground truth normals.
    }
    \small
    \centering 
    \setlength\tabcolsep{0pt}
    \begin{tabular*}{0.98\columnwidth}{@{\extracolsep{\fill}} ccccccccc }
    \hline
    
    \multirow{2}[2]{*}{Methods} & \multicolumn{3}{c}{\tabincell{c}{Rendered}} & \multicolumn{3}{c}{\tabincell{c}{Albedo}} & Normal \\
    
    \cmidrule{2-4} \cmidrule{5-7} \cmidrule{8-8} 
    &\tabincell{c}{PSNR $\uparrow$}     &\tabincell{c}{SSIM $\uparrow$}  &\tabincell{c}{MSE $\downarrow$}  
    &\tabincell{c}{PSNR $\uparrow$}     &\tabincell{c}{SSIM $\uparrow$}  &\tabincell{c}{MSE $\downarrow$} 
    &\tabincell{c}{MAE(${}^{\circ}$)  $\downarrow$} \\
    
    \hline
            
        NeRF-OSR   &22.75 &0.808 &0.009 &20.16 &0.825 &0.014 &26.21 \\
        
        SOL-NeRF   &23.44 &0.863 &0.006 &24.89 &0.827 &0.006 &16.33 \\

        ReCap      &26.34 &0.919 &0.004 &17.57 &0.774 &0.028 &30.55 \\

        LumiGauss  &24.97 &0.897 &0.004 &24.30 &0.884 &0.005 &33.88 \\
        
        Ours      & \textbf{26.75} & \textbf{0.933} & \textbf{0.003} 
                  & \textbf{26.80} & \textbf{0.869} & \textbf{0.003} 
                  & \textbf{13.97} \\

    \hline
    \end{tabular*}
    \label{tab:scene_decomp}
\end{table}

%% file: fig_decomp.tex
\begin{figure*}[tp]
    \centering
    {
    \setlength\tabcolsep{0pt}
    \includegraphics[width=\linewidth]{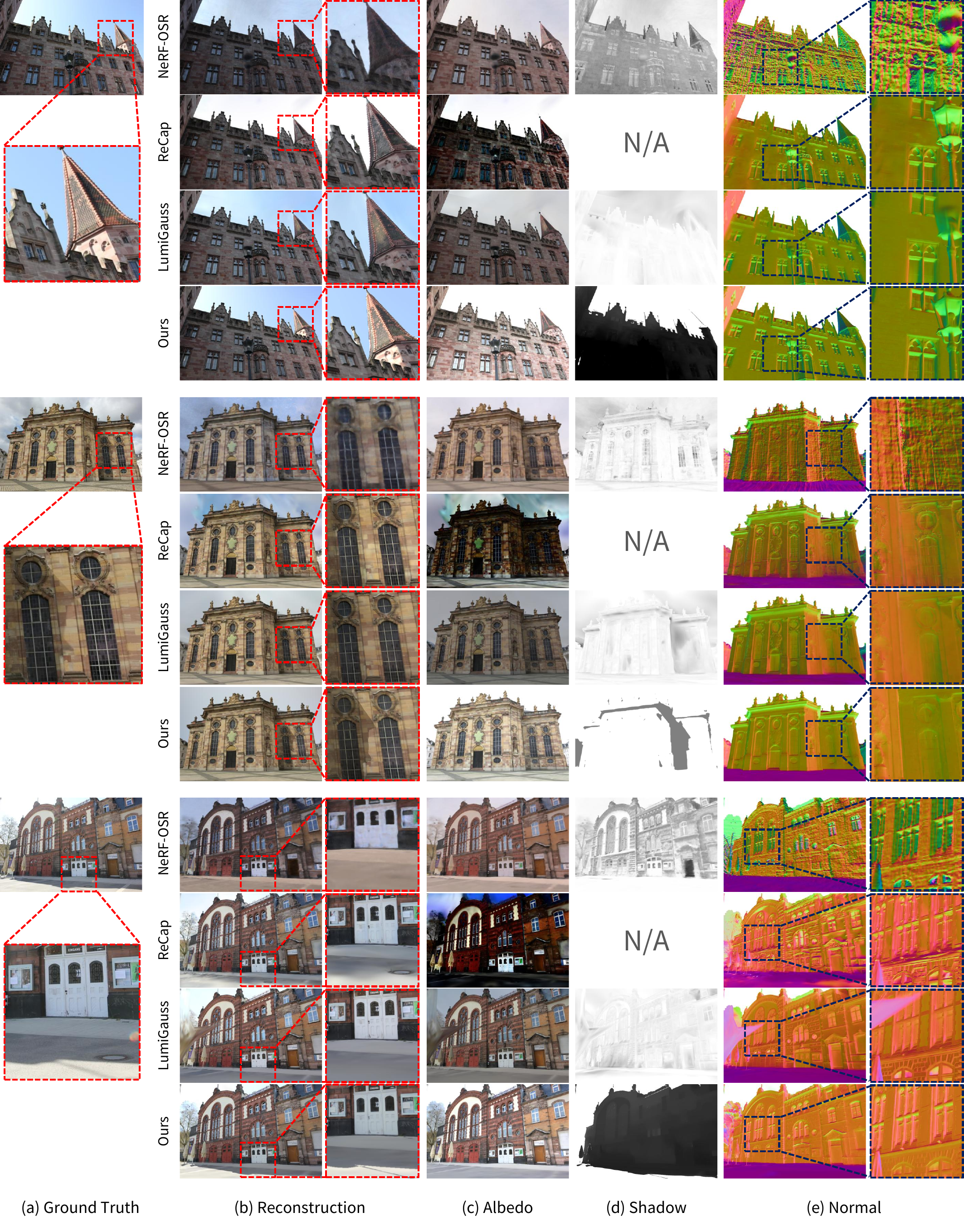}

    }
    \caption{
        \textbf{Qualitative comparisons of decomposition results} on baseline methods and our method. For each scene, we show decomposed components (normal, albedo, and shadow) and the reconstructed image. 
    }
    \label{fig:decomp}
\end{figure*}

%% file: fig_decomp_extra.tex
\begin{figure*}[t!]
    \centering
    \setlength\tabcolsep{0pt}
    \includegraphics[width=\linewidth]{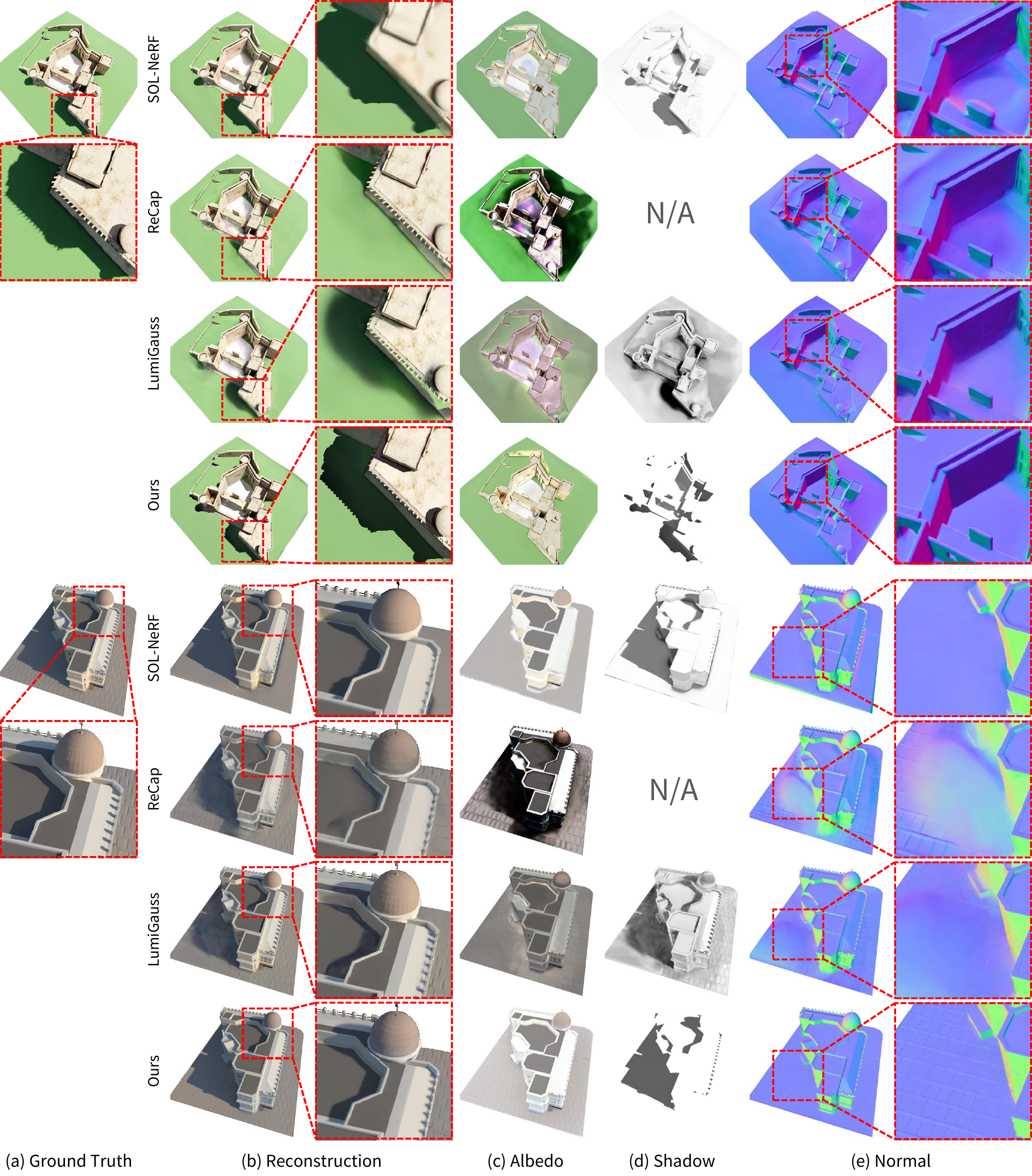}
    \caption{
        \textbf{Qualitative comparison of decomposition results on synthetic scenes}.
        We present reconstruction, albedo, shadow and normal results of our method compared with SOL-NeRF~\cite{solnerf}, ReCap~\cite{ReCap} and LumiGauss~\cite{LumiGauss} on two synthetic scenes.
    }
    \label{fig:decomp_extra}
\end{figure*}

%% file: fig_decomp_neusky.tex
\begin{figure*}[t!]
    \centering
    \setlength\tabcolsep{0pt}
    \includegraphics[width=\linewidth]{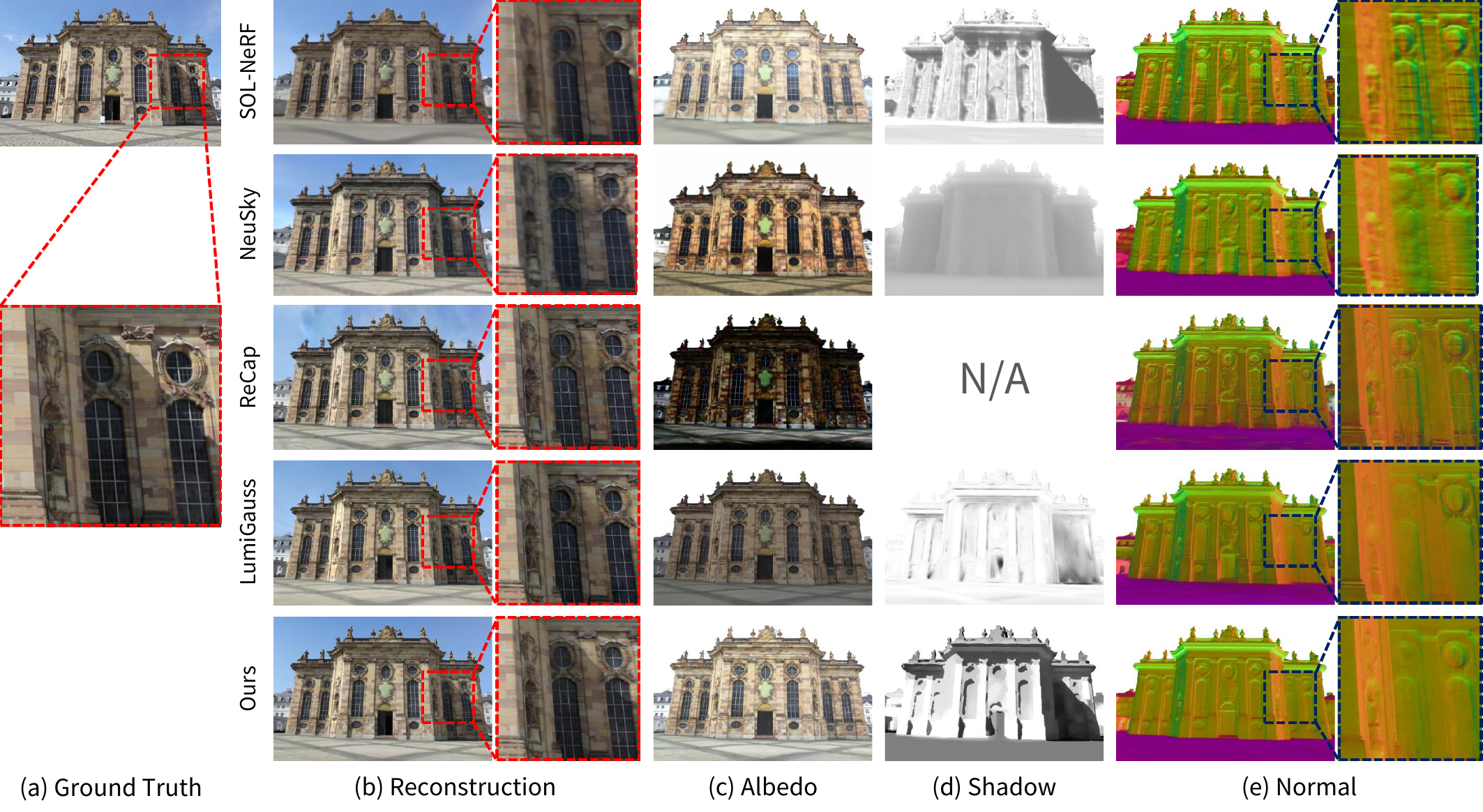}
    \caption{
        \textbf{Additional qualitative comparison of decomposition results}.
        We present reconstruction, albedo, shadow and normal results of our method compared with SOL-NeRF~\cite{solnerf}, NeuSky~\cite{NeuSky}, ReCap~\cite{ReCap} and LumiGauss~\cite{LumiGauss} on a real scene.
    }
    \label{fig:decomp_neusky}
\end{figure*}

%% file: table_relight.tex
\begin{table*}[t!]
    \caption{
        \textbf{Quantitative comparison of relighting results} using PSNR, SSIM, MAE and MSE metrics on real scenes. 
        $u/s$ denotes using up-sampled images for evaluation.
        We use environment maps provided by the NeRF-OSR dataset. 
        Results are averaged over five novel views in the test set.
    }

    \small
    \centering
    \setlength\tabcolsep{0pt}
    \begin{tabular*}{0.96\textwidth}{@{\extracolsep{\fill}} lcccccccccccc }
    
    \hline
    
    \multirow{2}[2]{*}{Methods} 
        & \multicolumn{4}{c}{\tabincell{c}{Ludwigskirche}} 
        & \multicolumn{4}{c}{\tabincell{c}{Staatstheater}} 
        & \multicolumn{4}{c}{\tabincell{c}{Landwehrplatz}}
        \\
    \cmidrule{2-5} \cmidrule{6-9} \cmidrule{10-13}
    &\tabincell{c}{PSNR $\uparrow$} &\tabincell{c}{MSE $\downarrow$} &\tabincell{c}{MAE $\downarrow$} &\tabincell{c}{SSIM $\uparrow$}
    &\tabincell{c}{PSNR $\uparrow$} &\tabincell{c}{MSE $\downarrow$} &\tabincell{c}{MAE $\downarrow$} &\tabincell{c}{SSIM $\uparrow$}
    &\tabincell{c}{PSNR $\uparrow$} &\tabincell{c}{MSE $\downarrow$} &\tabincell{c}{MAE $\downarrow$} &\tabincell{c}{SSIM $\uparrow$}
    \\
    
    \hline
    Yu et al.$_{u/s}$~\cite{ECCV2020Yu} &17.87 &0.017 &0.097 &0.378 &15.28 &0.032 &0.138 &0.385 &15.17 &0.033 &0.133 &0.376 \\

    Philip et al.~\cite{TOGPhilip19} &16.63 &0.023 &0.113 &0.367 &12.34 &0.065 &0.200 &0.272 &12.28 &0.062 &0.179 &0.319 \\

    NeRF-OSR~\cite{NeRF-OSR}    &18.72 &0.014 &0.090 &0.468 &15.43 &0.029 &0.133 &0.517 &16.65 &0.024 &0.114 &0.501 \\

    FEGR~\cite{FEGR} &21.53 &0.007 &- &- &17.00 &0.023 &- &- &17.57 &0.018 &- &- \\

    SOL-NeRF~\cite{solnerf}    &21.23 &0.008 &- &0.749 &18.18 &0.019 &- &0.680 &17.58 &0.028 &- &0.618 \\

    SR-TensoRF~\cite{SRTensoRF} &17.30 &0.021 &0.096 &0.542 &15.43 &0.030 &0.111 &0.632 &16.74 &0.024 &0.093 &0.653 \\

    NeuSky~\cite{NeuSky} &22.50 &0.005 &- &- &16.66 &0.023 &- &- &18.31 &0.016 &- &- \\

    LumiGauss~\cite{LumiGauss} &19.59  &0.012  &0.085  &0.700  &17.02  &0.021  &0.107  &0.729  &18.01  &0.017  &0.096  &0.778 \\

    ReCap~\cite{ReCap}    &24.11  &0.004  &0.041  &0.808  &22.70  &0.006  &0.051  &0.819  &21.55  &0.008  &0.051  &0.837 \\

    Ours        &\textbf{25.33} &\textbf{0.003} &\textbf{0.027} &\textbf{0.833} 
                &\textbf{23.62} &\textbf{0.005} &\textbf{0.036} &\textbf{0.833}
                &\textbf{21.87} &\textbf{0.007} &\textbf{0.043} &\textbf{0.847} 
                \\

    \hline
    \end{tabular*}
    \label{tab:relight}
\end{table*}

%% file: fig_relight.tex
\begin{figure*}[tp]
    \centering

    \includegraphics[width=\textwidth]{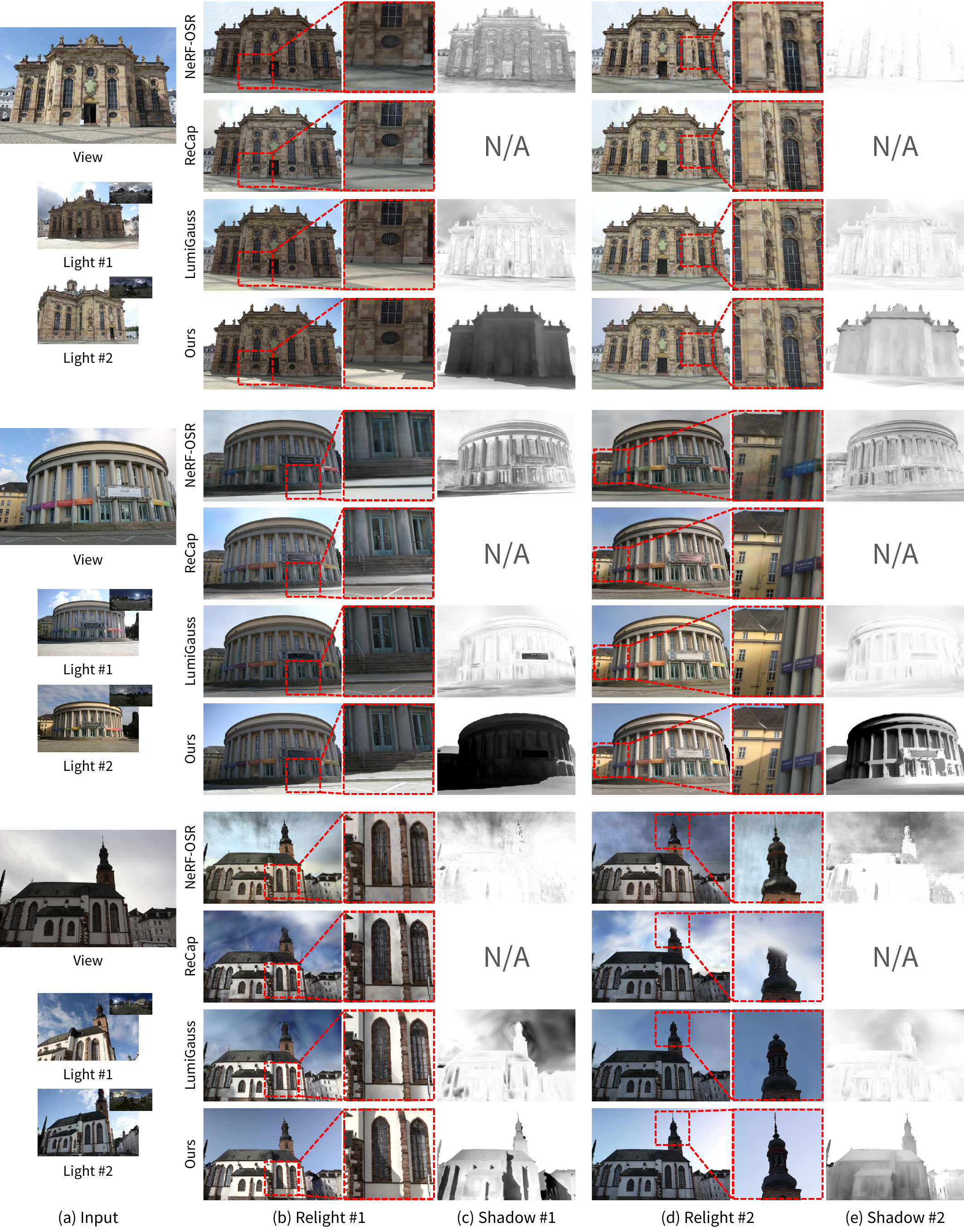}

    \caption{
        \textbf{Qualitative comparisons of relighting results} on baseline methods and our method. 
        For each input view, we relight it with two different lighting conditions and show rendered images and shadow maps. 
    }
    \label{fig:relighting}
\end{figure*}

%% file: fig_shadow_moving.tex
\begin{figure*}[tp]
    \centering
    \setlength\tabcolsep{0pt}
    \includegraphics[width=\linewidth]{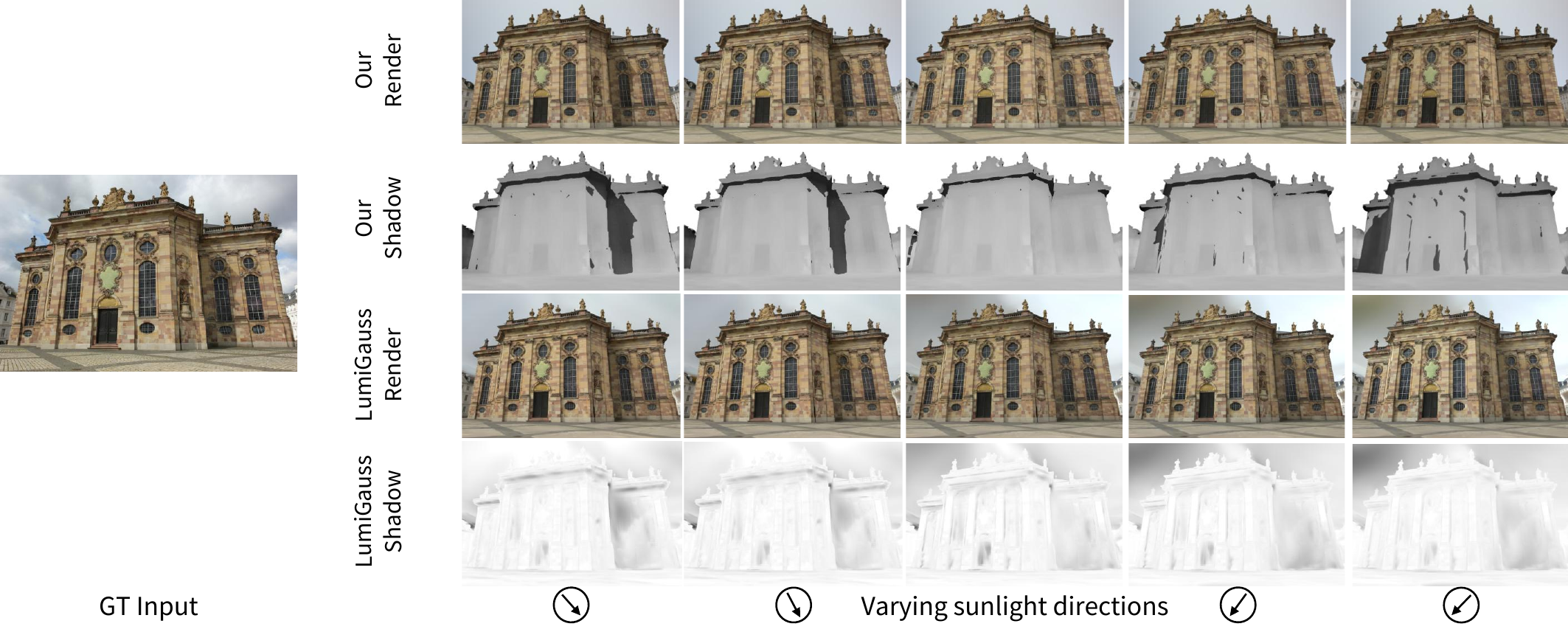}
    \caption{
        \textbf{Qualitative comparison of shadow movement effects} under varying sunlight directions between our method and LumiGauss~\cite{LumiGauss}. 
        For LumiGauss, relighting results are generated by rotating its SH lighting representation.
    }
    \label{fig:shadow_moving}
\end{figure*}

%% file: fig_ablation_L_np.tex
\begin{figure}[t!]
    \centering
    {
    \includegraphics[width=\linewidth]{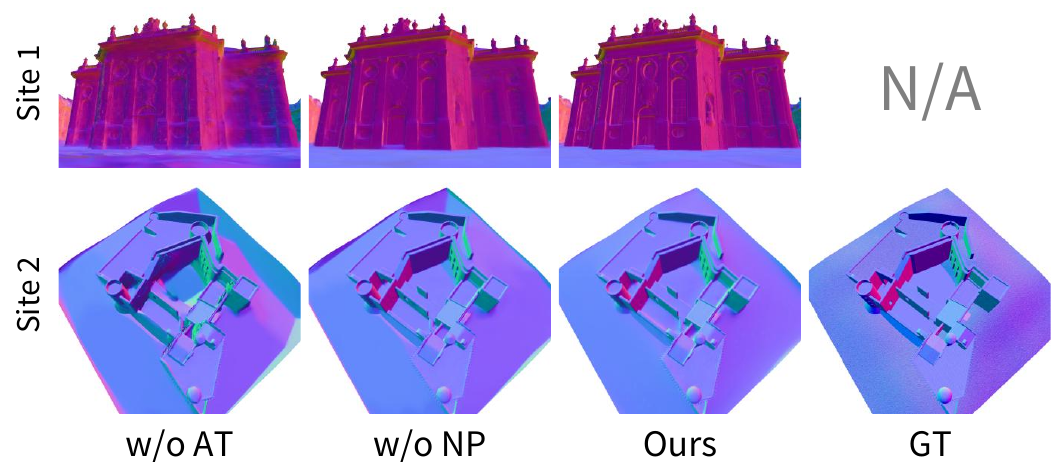}
    }
    \vspace{-3mm}
    \caption{
        \textbf{Qualitative comparison} of decomposed normal results with two variants.
        w/o AT refers to the variant that does not incorporate appearance transformation module. 
        w/o NP refers to the variant that does not apply monocular normal estimation as prior.
        Note that ground truth is unavailable for real data. 
    }
    \label{fig:ablation_L_np}
\end{figure}

%% file: table_efficiency.tex
\begin{table}[t!]
    \caption{
        \textbf{Performance comparison} with baselines.
    }
    \small
    \centering
    \setlength{\tabcolsep}{6pt}
    \renewcommand{\arraystretch}{1.1}
    \begin{tabular}{lcc}
    \hline
    Method      & Training Time & FPS \\
    \hline
    NeRF-OSR    & 31h           & 0.003 \\
    ReCap       & 2.7h          & 71.1  \\
    LumiGauss   & 1.3h          & 29.6  \\
    Ours        & 2.9h          & 38.6  \\
    \hline
    Ours SH$\times$AO  &3.8h  &10.6  \\
    \hline
    \end{tabular}
    \label{tab:efficiency}
\end{table}

%% file: table_ablation.tex
\begin{table}[t!]
    \caption{
        \textbf{Quantitative comparison of ablation study results} on the synthetic dataset. 
        \textit{w/o AT}: removal of the appearance transformation module; 
        \textit{w/o NP}: removal of the normal prior loss $\mathcal{L}_{np}$; 
        \textit{SH$\times$AO}: SH-based ambient light with MLP-predicted AO;
        \textit{w/o SG}: removal of spherical Gaussian (SG) sunlight modeling; 
        \textit{Forward}: forward shading; 
        \textit{Joint}: joint training instead of our two-stage strategy; 
        \textit{Ours}: our full model.
    }
    \small
    \centering 
    \setlength\tabcolsep{0pt}
    \begin{tabular*}{0.98\columnwidth}{@{\extracolsep{\fill}} ccccccccc }
        \hline
        \multirow{2}{*}{Methods} & \multicolumn{3}{c}{Rendered} & \multicolumn{3}{c}{Albedo} & Normal \\
        \cmidrule(lr){2-4} \cmidrule(lr){5-7} \cmidrule(lr){8-8}
        & PSNR $\uparrow$ & SSIM $\uparrow$ & MSE $\downarrow$
        & PSNR $\uparrow$ & SSIM $\uparrow$ & MSE $\downarrow$
        & MAE (${}^{\circ}$) $\downarrow$ \\
        \hline
        w/o AT    & 19.14 & 0.753 & 0.012 & 19.19 & 0.754 & 0.012 & 47.99 \\
        w/o NP    & 20.84 & 0.763 & 0.009 & 20.78 & 0.758 & 0.009 & 36.05 \\
        SH$\times$AO  & 20.33 & 0.905 & 0.014 & 20.37 & 0.729 & 0.010 & 13.97 \\
        w/o SG    & 23.87 & 0.907 & 0.005 & 21.89 & 0.756 & 0.007 & 13.97 \\
        Forward   & 23.32 & 0.877 & 0.005 & 20.27 & 0.722 & 0.009 & 13.97 \\
        Joint     & 20.76 & 0.791 & 0.010 & 20.37 & 0.729 & 0.010 & 21.84 \\
        Ours      & \textbf{26.75} & \textbf{0.933} & \textbf{0.003} 
                  & \textbf{26.80} & \textbf{0.869} & \textbf{0.003} 
                  & \textbf{13.97} \\
        \hline
    \end{tabular*}
    \label{tab:ablation}
\end{table}

%% file: fig_ablation_noSG.tex
\begin{figure}[t!]
    \centering
    {
    \includegraphics[width=\linewidth]{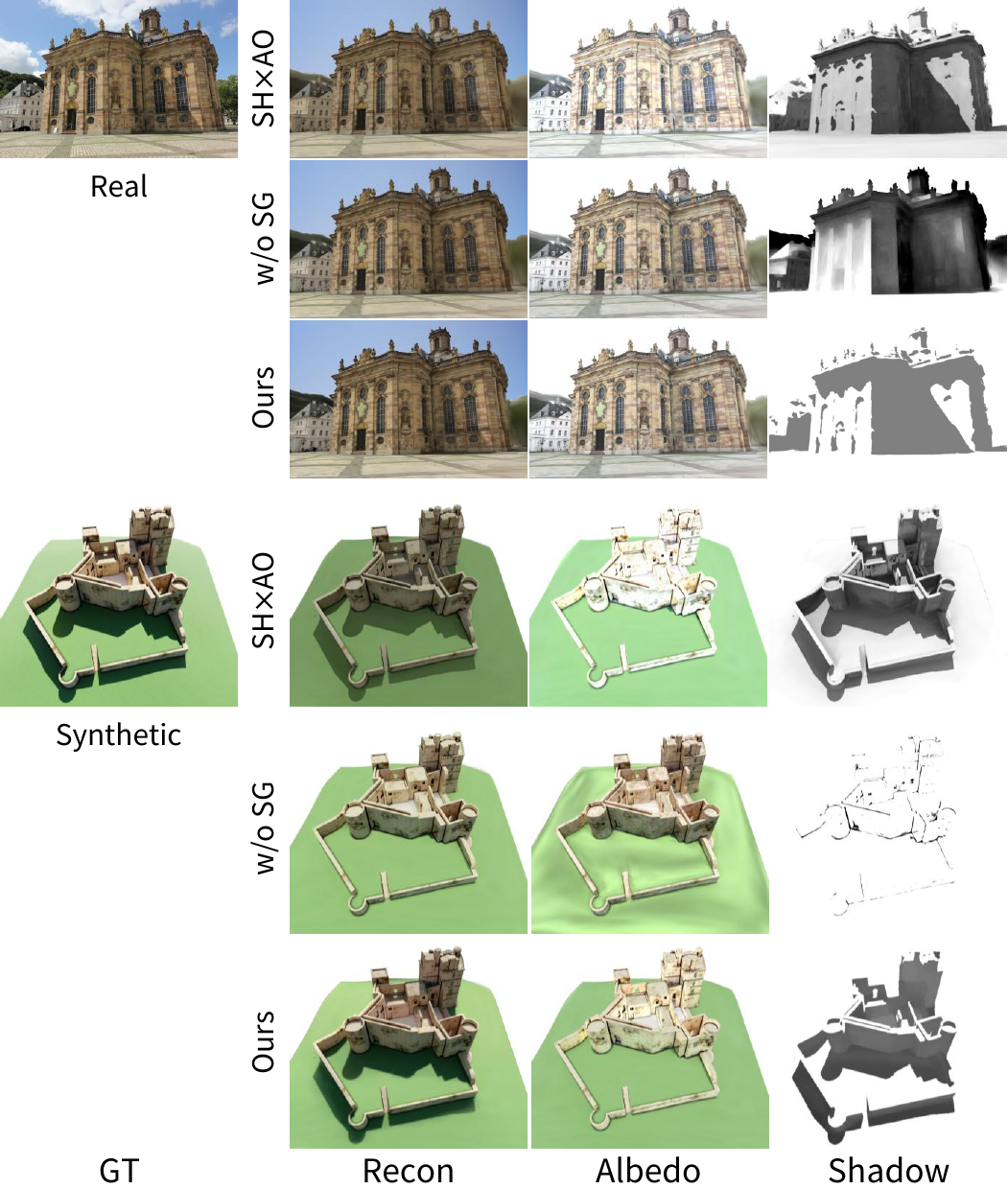}
    }
    \caption{
        \textbf{Qualitative comparison} of reconstruction results between our full model and two variants.
        \textit{SH$\times$AO} denotes the variant adopting the same lighting model as SOL-NeRF~\cite{solnerf}, which uses an SH-based environment map with an MLP-predicted ambient occlusion.
        \textit{w/o SG} indicates the variant without Spherical Gaussian (SG) sunlight modeling, leading to inaccurate shadows and degraded albedo quality.
    }
    \label{fig:ablation_noSG}
\end{figure}

%% file: fig_ablation_joint.tex
\begin{figure}[th]
    \centering
    {
    \includegraphics[width=\linewidth]{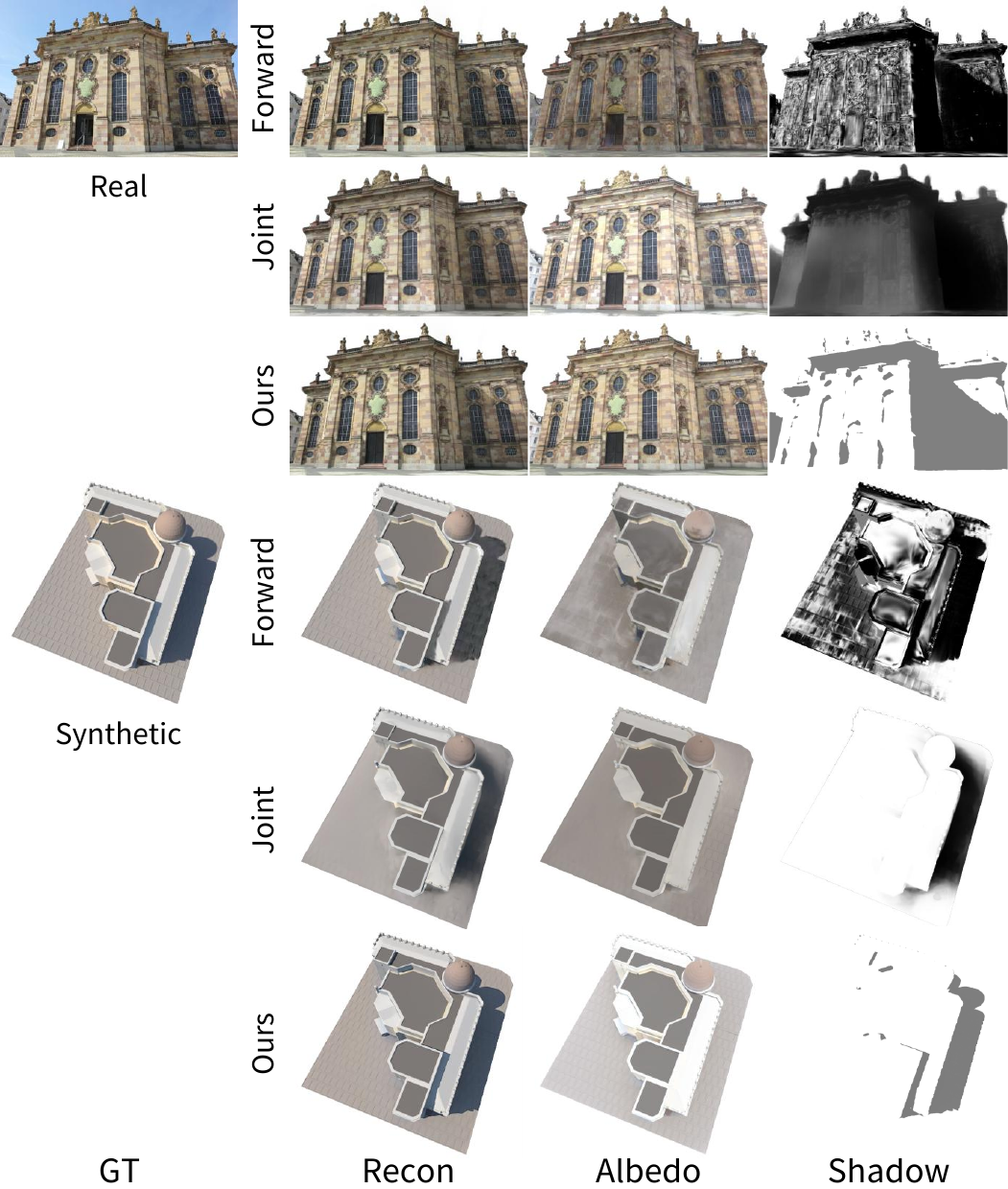}
    }
    \vspace{-3mm}
    \caption{
        \textbf{Qualitative comparison} of decomposed diffuse albedo and shadow maps with two variants.
        "Forward" denotes a variant based on forward shading, while "Joint" indicates a joint training strategy in contrast to our two-stage approach.
        Note that our two-stage method significantly improves reconstruction and shadow separation.
    }
    \label{fig:ablation_joint}
\end{figure}

%% file: fig_failure_case.tex
\begin{figure}[t]
    \centering
    {
    \includegraphics[width=\linewidth]{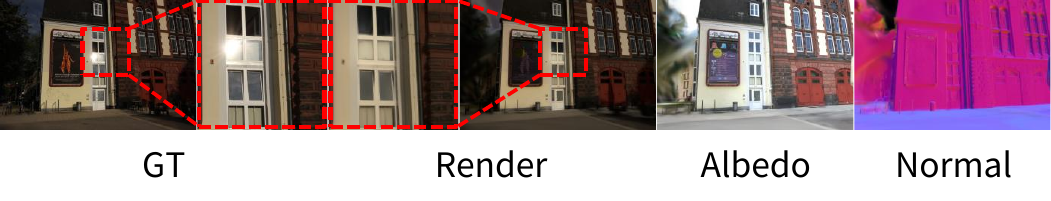}
    }
    \vspace{-3mm}
    \caption{
        \textbf{Qualitative comparison} of a failure case with a reflective surface exhibiting strong specular highlights.
    }
    \label{fig:failure_case}
\end{figure}

%% file: 5_conclusion.tex
\section{Conclusion}
\subsection{Technical Summary}
We introduced ROS-GS, a Gaussian Splatting framework for outdoor relighting, effectively combining rapid rendering with physically-based realism for photorealistic lighting manipulation. Its two-stage approach first reconstructs scene geometry using 2DGS and monocular normal priors from multi-view images. Subsequently, it employs physically-based rendering and a hybrid sun-sky lighting model to decompose view-independent texture, high-frequency sunlight, and low-frequency skylight. Progressive optimization allows ROS-GS to disentangle these components from images captured under varying conditions, enabling plausible, efficient outdoor relighting. Experiments demonstrate ROS-GS surpasses current NeRF-based and 3DGS-based baselines.

\subsection{Limitations and Future works}
Our current model assumes primarily diffuse surface materials, which may be suboptimal for highly reflective scenes as shown in Fig.~\ref{fig:failure_case}. Additionally, its reliance on mesh-based ray tracing for visibility omits light interactions with dynamic scene elements. Future work will address these by incorporating more sophisticated material models and exploring advanced rendering techniques to handle complex illumination and dynamic scenes more robustly, thereby enhancing realism and applicability.